\documentclass[10pt,twocolumn,letterpaper]{article}

\usepackage{cvpr}
\usepackage{times}
\usepackage{epsfig}
\usepackage{graphicx}
\usepackage{amsmath}
\usepackage{amssymb}
\usepackage{booktabs,subcaption,amsfonts,dcolumn}
\usepackage{imakeidx}
\makeindex

\newcolumntype{d}[1]{D..{#1}}
\usepackage{rotating}
\usepackage{tabularx}
\usepackage[table]{xcolor}
\usepackage[pagebackref=true,breaklinks=true,letterpaper=true,colorlinks,bookmarks=false]{hyperref}

\cvprfinalcopy 

\def\cvprPaperID{****} 

\newcommand{\eman}[1]{\ifbool{inccomment}{{\color{blue} #1}}{}}
\newcommand{\xin}[1]{\ifbool{inccomment}{{\color{blue} #1}}{}}
\newcommand{\david}[1]{\ifbool{inccomment}{{\color{red} #1}}{}}

\ifcvprfinal\fi

\begin{document}

\title{Unsupervised Domain Adaptation using Generative Models and Self-ensembling}

\author{Eman T. Hassan \\
Indiana University Bloomington\\
{\tt\small emhassan@indiana.edu}
\and
Xin Chen \\
Midea ETC\\
{\tt\small chen1.xin@midea.com }
\and
David Crandall \\
Indiana University Bloomington\\
{\tt\small djcran@indiana.edu}
}

\maketitle

\begin{abstract}
Transferring knowledge across different datasets is an important approach to successfully train deep models with a small-scale target dataset or when few labeled instances are available. In this paper, we aim at developing a model that can generalize across multiple domain shifts, so that this model can adapt from a single source to multiple targets. This can be achieved by randomizing the generation of the data of various styles to mitigate the domain mismatch. First, we present a new adaptation to the CycleGAN model to produce stochastic style transfer between two image batches of different domains. Second, we enhance the classifier performance by using a self-ensembling technique with a teacher and student model to train on both original and generated data. Finally, we present experimental results on three datasets \textbf{Office-31}, \textbf{Office-Home}, and \textbf{Visual Domain adaptation}. The results suggest that self-ensembling is better than simple data augmentation with the newly generated data and a single model trained this way can have the best performance across all different transfer tasks.
\end{abstract}

\newcommand{\genCellVis}[3]{
\setlength\fboxsep{0pt}
\begin{minipage}{3.9cm}
\includegraphics[width=1.\textwidth]{#1} \\
\includegraphics[width=1.\textwidth]{#2} \\
\includegraphics[width=1.\textwidth]{#3} 
\end{minipage}}

\newcommand{\genRowVis}[6]{
\genCellVis{#1}{#3}{#5} \genCellVis{#2}{#4}{#6}
}

\newcommand{\genCellOffice}[3]{
\setlength\fboxsep{0pt}
\begin{minipage}{3.9cm}
\includegraphics[width=1.\textwidth]{off_#1} \\
\includegraphics[width=1.\textwidth]{off_#2} \\
\includegraphics[width=1.\textwidth]{off_#3} 
\end{minipage}}

\newcommand{\genRowOffice}[6]{
\genCellOffice{#1}{#3}{#5} \genCellOffice{#2}{#4}{#6}
}

\newcommand{\genCellAmazon}[3]{
\setlength\fboxsep{0pt}
\begin{minipage}{3.9cm}
\includegraphics[width=1.\textwidth]{amz_#1} \\
\includegraphics[width=1.\textwidth]{amz_#2} \\
\includegraphics[width=1.\textwidth]{amz_#3} 
\end{minipage}}

\newcommand{\genRowAmazon}[6]{
\genCellAmazon{#1}{#3}{#5} \genCellAmazon{#2}{#4}{#6}
}


\section{Introduction}
\begin{figure*}
\begin{center}
\includegraphics[width=.8\linewidth]{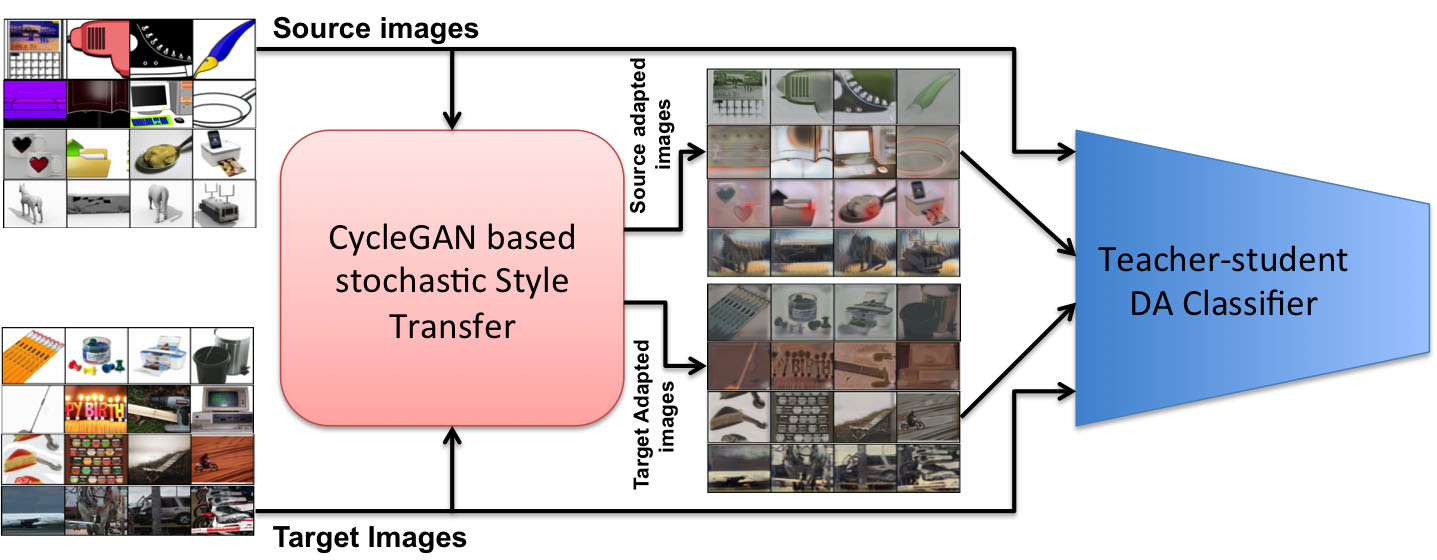}
\end{center}
\vspace{-20pt}
   \caption{The overall system architecture. The input source and target images are sampled randomly from \textbf{Office-31} \cite{ds_amazon}, \textbf{Office-Home} \cite{ds_officehome}, and \textbf{Visual Domain adaptation} \cite{ds_visda} datasets. The images are fed to the CycleGAN-based Stochastic Style Transfer module (explained in more detail in Figure \ref{fig:system_all}). Then style adapted images are generated for both source and target domains. All the images, source and target (both original and adapted), are fed to a teacher-student DA classifier system (explained in more detail in Figure \ref{fig:semimodel}).}
\label{fig:system_bigPict}
\end{figure*}

Large scale annotated dataset and increasing computational power have enabled the rapid development of deep convolutional networks~(CNNs) that produce high performance on many computer vision problems like classification, segmentation, and detection~\cite{alexNetClass,segm_intro,faster_rcnn}. Unlike human beings who easily generalize knowledge across different domains, it is not easy for CNN models to generalize across other datasets having different characteristics.
As a result, domain adaptation, which aims at enhancing the ability of the network to generalize across different domains, has emerged as a hot topic in recent years.



Training a generative model to generate images with different characteristics has witnessed significant progress. Especially after the introduction of Generative Adversarial Networks (GAN)~\cite{GAN_intro}. 
GANs enable the development of deep models which are capable of image generation across different domains with high quality \cite{intro_dcgan,intro_stackedGan,intro_progressiveGAN}. Therefore, GANs have been widely applied to many areas such as image inpainting~\cite{intro_encoder}, super-resolution~\cite{intro_superRes}, pixel-level domain adaptation~\cite{adver_pixels}, and image-to-image translation~\cite{cycleGAN,cycleGAN_augment}.

That has inspired us to use GANs to randomize the generation of multiple instances of the dataset of different styles to accommodate for any possible unseen domain shifts. We then use these newly generated datasets to effectively enhance knowledge transfer in a teacher-student framework. Previous work~\cite{cycleGAN_atten_DA,rec_seperationNet} has examined using generative networks as a part of the overall adaptation between two domains, while in our case we examine the possibility of generative networks to generate many random instances of the original datasets, where each instance can represent a specific domain shift. Finally we employ the newly obtained datasets to enhance zero-shot domain adaptation.

In this paper, we propose a model that can generalize across different domain shifts. This is accomplished in two steps, the overall proposed system is shown in Figure~\ref{fig:system_bigPict}.. First, we introduce stochastic style transfer as an adaptation to the CycleGAN network \cite{cycleGAN}. It transforms the function of the module from performing one-to-one image translation, to instead perform stochastic style transfer, which generates images with adapted style between the source and target domains. The adaptation depends on relaxing some of the mapping constraints between the two domains. It only checks the mapping between the two domains based on a disentangled representation of the images into style and content representations. Second, we use the trained modules to generate newly adapted datasets. Then, to achieve zero-shot domain adaptation, we develop self-ensembling based domain adaptation, which adapts the teacher-student architecture proposed in~\cite{semi_DA_imprt}, where different instances of the same data are the source of perturbation. The experimental results suggest that the newly proposed architecture produces better performance on transferring between one domain to many others.

In summary,  we make the following major contributions in this paper:
\begin{enumerate}
    \item Showing that GAN networks can help make a model generalize across multiple domain shifts simultaneously. 
    \item Using CyleGAN-based architecture to perform stochastic style transfer between the two domains, which helps generate multiple instance of the original datasets.
    \item Employing a self-ensembling technique to train a model both in supervised and unsupervised ways using the original datasets and the newly generated instances for both source and target images.
    \item Showing that the self-ensembling architecture is better in training than fine tuning with the newly generated data.
\end{enumerate}

\section{Related Work}
\textbf{Domain adaptation} is a machine learning problem whose goal is that a model trained on a source data $D_s(I_s,Y_s)$ can generalize to related but different data $D_t(I_t,Y_t)$. The source data is fully labelled while the target data is either partially labeled (semi-supervised case) or totally unlabeled (unsupervised case). With this paradigm, many both shallow and deep techniques were proposed to tackle this problem. There are examples of \textit{\textbf{Shallow Techniques}} like Sup-space alignment \cite{sallow_sa}, second order statistic alignment known as CORAL \cite{shallow_coral}, the employment of landmarks to enhance feature alignment \cite{shallow_landmarksRahaf,shallow_landmarksGong,shallow_landmarkmshsa}, sample reweighting \cite{shallow_rewightbasics}, and Metric based learning as in \cite{shallow_metric_1,shallow_metric_2,shallow_metric_3}. Some examples of \textit{\textbf{Deep domain adaptation}} are knowledge distillation with soft-targets \cite{softTargets} and selectively choosing data samples for fine-tuning \cite{sampl_fineSample}. Generally, deep techniques can be categorized into discrepancy based methods, reconstruction based techniques and adversarial based techniques \cite{survey_1}. 

\textit{\textbf{Discrepancy based methods}} as in \cite{discrep_DAN,discrep_JAN,discrep_WJAN,discrep_DCORAL}  aim at alignment the deep feature embedding between the source and target domains by optimizing a loss function that penalizes distribution mismatch. Long \etal \cite{discrep_DAN} proposed a domain adaptation network that uses a maximum mean discrepancy loss function to match the distribution of the fully connected layers of the source and target networks, while Long \etal \cite{discrep_JAN} extended this work by using a joint maximum mean discrepancy loss function instead, and Yan \etal \cite{discrep_WJAN} incorporated the class weights in the loss function by employing an expectation maximization training paradigm. Similar to CORAL \cite{shallow_coral} in shallow techniques, Sun \etal \cite{discrep_DCORAL} used Deep CORAL to match the second order statistics between the deep features of the two networks. 

\textit{\textbf{Adversarial based techniques}} \cite{adver_gru,adver_coupledGAN,adver_disDA,adver_pixels} use adversarial settings to make the deep feature embeddings of both domains similar. Ganin \etal \cite{adver_gru} introduced gradient reverse units that reverse the gradient of the domain classifier so that the network learns similar feature representations for both domains. Tzeng \etal \cite{DomainConfus} incorporated loss functions like domain confusion and domain classification in the network framework. Tzeng \etal \cite{adver_disDA} proposed a general framework for unsupervised domain adaptation in both cases of generative or discriminative mappings. Bousmalis \etal \cite{adver_pixels} introduced a technique that adapts source-domain images to appear as if drawn from the target domain. \textit{\textbf{Reconstruction based techniques}} outline how to construct a latent representation that is shared across multiple domains. Ghifary \etal \cite{rec_visAutoEncoders} considered inter-domain variances as sources of noise in a denoising encoder, then make it learn common features across these different domains. Similarly Ghifary \etal \cite{rec_autoEnco2} proposed a deep reconstruction network that learns both labeled image prediction and the reconstruction of both target and source images. Bousmalis \etal \cite{rec_seperationNet} proposed a separation network that learns to separate the latent space into two components, one  private to each domain and the other  common across different domains. Peng \etal \cite{style_corr} proposed a deep generative alignment network that reconstruct images from the source domain that are similar to the target domain and trains the classifier based on these generated images. In our work, we assume a zero-shot framework where we penalize the difference on prediction of the target data, without actually trying to match the distribution of the source domain.

\begin{figure*}[h]
\begin{center}
\includegraphics[width=.8\linewidth]{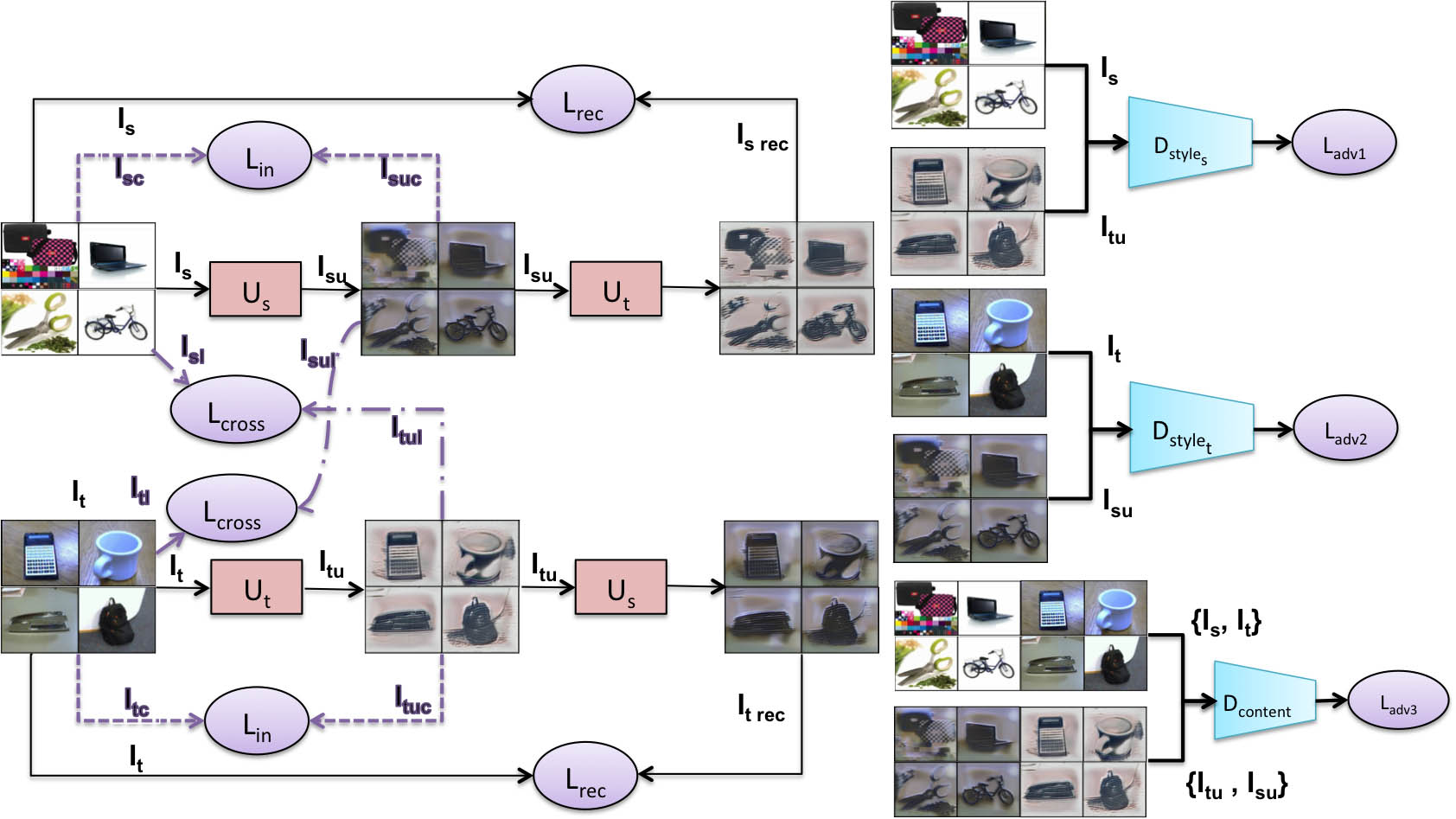}
\end{center}
   \vspace{-15pt}
   \caption{The CycleGAN-based Stochastic Style Transfer module, where input source and target images (sampled randomly from the \textbf{Office-31} dataset \cite{ds_amazon}) are fed to source $U_s$ and target $U_t$ Unets respectively. $U_s$ and $U_t$ nets produce images that match the content of the same domain and the style of the other domain. The resulting images are fed again but to cross domain Unets $U_t$ and $U_s$ respectively to get images that should be similar to the original input images in both content and style. Finally the discriminator networks $D_{Style_s}$ and $D_{Style_t}$ aim to confuse the generator regarding the style of the source and target domain respectively, while the discriminator network $D_{content}$ aims at fooling the generators to generate realistic images.}
\label{fig:system_all}
\end{figure*}

\textbf{Image-to-Image Translation} techniques describe the generation of an image that represents a translation of the input image from one domain to another~\cite{ImToIm_condGAN,cycleGAN,cycleGAN_augment,cycle_Multi_MToM,cycleGAN_atten_DA,ImToIm_adapt1,ImToIm_encoder}. Liu \etal \cite{ImToIm_encoder} proposed a UNIT framework for image translation that is based on generative adversarial networks and variational autoencoders that creates a shared latent space which can generate corresponding images between the two domains. Zhu \etal \cite{cycleGAN} presented the Cycle GAN model which introduced cycle consistency loss to construct a mapping between the two domains, but this model constructs a one-to-one mapping between the two domain. This work was followed by many papers that introduces models that make the translation many-to-many, as in \cite{cycle_Multi_MToM} where  Zhu \etal introduced a hybrid model that combines  conditional variational autoencoder-GANs and Conditional Latent Regressor-GANs. Kang \etal~\cite{cycleGAN_atten_DA} used an image generation model to enhance domain adaptation and made the source  network  guide the  attention  alignment  of  the target  network in an expectation maximization framework. In our work we extended CyleGAN to generate multiple style adaptations to the original domain to help in knowledge transfer.

\textbf{Semi-supervised learning and domain adaptation using teacher student model.}
Zagoruyko \etal \cite{semi_attention} showed that the performance of a student network is enhanced when it mimics the attention mechanism of the teacher network. Kang \etal \cite{cycleGAN_atten_DA} has employed deep adversarial attention between the two networks to achieve knowledge transfer between the two domains. Mean-teacher model \cite{semi_meanTeacher} and self-ensembling \cite{semi_temporal} were proposed in the framework of semi-supervised learning. Laine \etal \cite{semi_temporal} employed  the consensus in prediction of the network in training to enhance network performance. While Tarvainen \etal \cite{semi_meanTeacher} proposed the mean teacher model to aggregate information after each step not epoch, in this model the teacher weights used exponential moving average weights of the student model. French \etal \cite{semi_DA_imprt}  extended these ideas to the problem of domain adaptation. Meanwhile Luo \etal \cite{semi_neighbors_smooth} consider the connection between different data points to build a graph that employs a self-ensemble based on smooth neighbors on teacher graphs.

\section{Algorithm}
\subsection{CycleGAN based stochastic Style Transfer}
The motivation of this adaptation is how to generate images that have the same contents of the source domain but randomized style that can match in some aspect different domain shifts. The CycleGAN module \cite{cycleGAN} constructs a mapping between the two domains $X,Y$ as $F:X\longrightarrow Y$, $G:Y\longrightarrow X$ where the cycle consistency aims at producing $X \simeq G(F(X))$. That module is designed to produce a one-to-one mapping between the two domains $X,Y$. We propose a modification to the architecture so that the image generated does not belong to either $X$ or $Y$, but contains a shared style representation across the two domains. This architecture is shown in Figure \ref{fig:system_all}. The architecture assumes we can obtain a disentangled representation for the images into both content and style representations, so that each domain generator can produce an image that matches the content representation of the original domain and matches random style representation of the other domain. In this way we can benefit from the module to generate multiple random dataset instances that represent different shifts with respect to the original domain.

To get a disentangled feature representation for image $I$ into content $I_c$ and style $I_l$ representation, we employ the method in \cite{style_transVgg}. An input image $I_s$ is passed through a pre-trained Vgg-16 model\cite{vgg_trainednet} where $I_s=relu_{52}(I))$ and $I_t$ is the concatenation of $Gm(relu_{12}(I))$, $Gm(relu_{22}(I))$, $Gm(relu_{33}(I))$, $Gm(relu_{43}(I))$, and  $Gm(relu_{53}(I))$, where $Gm$ represents the gram matrix of the features of $relu_{ij}(I)$ represents the Rectified Linear Unit activation applied to the output of the $j^{th}$ convolution module in layer $i$.

In the proposed architecture $U_s$ and $U_t$ represent the source and target generator networks respectively which are implemented using the Unet model architecture \cite{unet_model}. Input images $I_s$ and $I_t$ from the source and target domains are forwarded to the system such that $I_{su} = U_s(I_s)$ and $I_{tu} = U_t(I_t)$. Then the content and style representations are extracted for both input and generated images. For example $I_{sc}$ and $I_{sl}$ are the content and style representations, respectively, of $I_s$, and similarly $I_{tuc}$ and $I_{tul}$ are the representations for $I_{tu}$. We introduce an Intra-domain loss function $l_{in}$ and cross domain loss function $l_{cross}$ to be able to generate images with similar contents of the original domain but similar style of the other domain. 
The Intra-domain loss function $l_{in}$ penalizes the difference in the content representation between the input and generated image of the same domain: 
\begin{equation} \label{eq:1}
    L_{in} = MSE(I_{sc},I_{suc})+MSE(I_{tc},I_{tuc})
\end{equation}
while the cross domain loss function $l_{cross}$ penalizes the difference in the style representation across the other domain: 
\begin{equation}\label{eq:2}
    L_{cross} = MSE(I_{sl},I_{tul}) + MSE(I_{tl},I_{sul})
\end{equation}
Then we have the cycle loss represented by $l_{recons}$. The reconstruction loss $l_{recons}$ contributes to training convergence and generating realistic images, and is computed by
\begin{flalign}\label{eq:3}
\begin{split}
    L_{rec} = MSE(I_{s},I_{s rec}) + MSE(I_t,I_{t rec}) \\
    \mbox{where,\,\,} I_{s rec} = U_t(U_s(I_s)) \\
    \mbox{where,\,\,} I_{t rec} = U_s(U_t(I_t)) 
\end{split}
\end{flalign}

Finally we calculate the adversarial loss functions $l_{adv1},l_{adv2}$ and $l_{adv3}$, respectively. Adversarial losses $l_{adv1},l_{adv2}$ are employed so that the discriminator networks $D_{style_s}$, $D_{style_t}$ confuse the domain generator networks $U_s$ and $U_t$ to generate images with style similar to the style of the other domain.
\begin{flalign}\label{eq:4}
\begin{split}
    l_{adv1} = \min_{\{U_s,U_t\}}\max_{D_{style_s} }\{E_{I_s}[\log(D_{style_s}(I_s) ) ] + \\
            E_{I_t}[\log(1-D_{style_s}( U_t(I_t) )  ) ] \}
\end{split}
\end{flalign}

\begin{flalign}\label{eq:5}
\begin{split}
    l_{adv2} = \min_{\{U_s,U_t\}}\max_{D_{style_t} }\{ E_{I_t}[\log(D_{style_t}(I_t) ) ] + \\
            E_{I_s}[\log(1-D_{style_t}( U_s(I_s) )  ) ] \}
\end{split}
\end{flalign}
On the other hand $l_{adv3}$ and $D_{content}$ confuse the generator networks $U_s$ and $U_t$ to generate realistic images. 
\begin{flalign}\label{eq:6}
\begin{split}
    L_{adv3} = \min_{\{U_s,U_t\}}\max_{D_{content}}\{ E_{\{I_s,I_t\}}[\log(D_{content}(\{I_s,I_t\}) )] + \\
            E_{\{I_s,I_t\}}[1-\log(D_{content}(\{U_s(I_s),U_t(I_t)\} ) ) ]\}
\end{split}
\end{flalign}
The final loss function is a weighted average of these losses, and is computed as: 
\begin{flalign}\label{eq:7}
\begin{split}
    l_{total} = \lambda_1\times l_{in} + \lambda_2\times l_{cross} + \lambda_3\times  l_{rec} + \\ \lambda_4\times  l_{adv1} + \lambda_5\times  l_{adv2} + \lambda_6\times l_{adv3}
\end{split}
\end{flalign}
\subsection{Self-ensemble zero shot domain adaptation}
\begin{figure}[!ht]
 \includegraphics[width=1.1\linewidth]{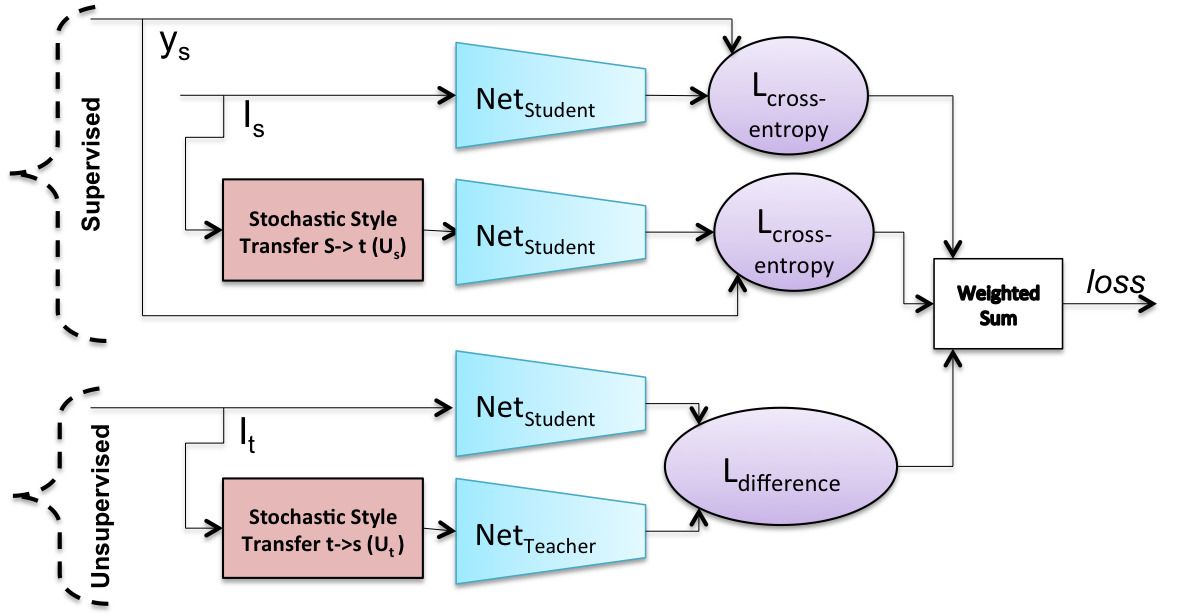}

\caption{The structure of the `Teacher-student' DA classifier mentioned in Figure \ref{fig:system_bigPict}.}
\label{fig:semimodel}
\end{figure}
The motivation of this section is that simple data augmentation may not be the best way to train a classifier benefiting from the generated images. As shown in Figure \ref{fig:system_bigPict} that we have both labelled source images and labelled randomly adapted source images and unlabelled target images and unlabelled randomly adapted target images, and data augmentation doesn't explore the dependencies that exists between each image and the corresponding adapted instance. 

The idea of a network to have consistent behaviour under different perturbations has been used to enhance the performance of the trained networks \cite{semi_peseudoEnsem,semi_ladderNet,semi_adverPerterb}. In our work we explore the idea that the perturbation can be adding a random new style to the image as shown in Figure \ref{fig:semimodel}. In this architecture we extend the framework provided by \cite{semi_DA_imprt} but we employ the perturbation to be stochastic style transfer between the two domains instead of random data augmentation. 

In this model the student  is trained in a supervised way using the source dataset instances (the original one $Data_{s}$ and the style mapped one $Data_{s_m}$). The student model is also trained in an unsupervised manner which is related to the difference in prediction between the student network and the teacher network to the original target data $Data_{t}$ and style mapped target $Data_{t_m}$. On the other hand, the teacher network model is updated in an exponential moving average manner with the student network. The supervised loss is described by: 
\begin{flalign}\label{eq:8}
\begin{split}
L_{sup}(\theta_{Student}) = -\frac{1}{N} \sum_{i=1}^{N}\{ \sum_{j=1}^{k}\{ 1\{y^{(i)}=j\} 
         \\
        \times \log\left( \frac{e^{\theta_{Student_j}^T x^{(i)}  } }{ \sum_{l=1}^{k}{ e^{\theta_{Student_l}^T x^{(i)}  } } }  \right)  \} \} 
\end{split}
\end{flalign}

The unsupervised loss is described by: 
\begin{equation}\label{eq:9}
    L_{unsup} = N_{Student}(I_{t}) - N_{Teacher}(I_{t_m}) 
\end{equation}

\section{Experimental Results}
We conducted experiments on three datasets for an image classification task:
\begin{enumerate}
    \item \textbf{Office-31} dataset \cite{ds_amazon}. It consists of 31 categories and contains $4,110$ images distributed over three domains: 1) \textbf{A}mazon domain which contains $2,817$ images that are collected from Amazon.com, 2) \textbf{W}ebcam domain containing $795$  web camera images, and 3) \textbf{D}slr domain containing 498 images from SLR cameras. In our work we evaluate the transfer of  $\textbf{A} \longrightarrow \{ \textbf{W} ,\textbf{D} \}$, $\textbf{W} \longrightarrow \{ \textbf{A} ,\textbf{D} \}$ and $\textbf{D} \longrightarrow \{ \textbf{W} ,\textbf{A} \}$.
    \item \textbf{Office-Home} dataset \cite{ds_officehome}. It consists of 65 categories in 15,590 images over four domains: 1) \textbf{Ar}t domain with 2427 images,  2) \textbf{Cl}ipart domain of 4,365 images downloaded from multiple clipart websites, 3) \textbf{P}roduct domain of 4,439 images  gathered from Amazon.com, and 4) \textbf{R}eal-world domain of 4,357 images. \textbf{Ar}t and \textbf{R}eal-World  domains were built from websites like www.deviantart.com and www.flickr.com. In our work we evaluate the transfer task $\textbf{Ar} \longrightarrow \{ \textbf{Cl},\textbf{P} ,\textbf{R} \}$, $\textbf{Cl} \longrightarrow \{ \textbf{Ar},\textbf{P} ,\textbf{R} \}$, $\textbf{P} \longrightarrow \{ \textbf{Cl},\textbf{Ar} ,\textbf{R} \}$ and $\textbf{R} \longrightarrow \{ \textbf{Cl},\textbf{P} ,\textbf{Ar} \}$.
    \item \textbf{Visual Domain Adaptation classification task (VisDa)} \cite{ds_visda}. It consists of over 280K images across 12 categories across two domains: 1) \textbf{Sy}nthetic and 2) \textbf{Re}al. In our work we evaluate the transfer between $\textbf{Sy}nthetic \longrightarrow \textbf{Re}al$.
\end{enumerate}

\begin{table*}[t]\centering
\begin{tabular}{c|c}
\textbf{Epoch 10} & \textbf{Epoch 15}  \\
\genRowOffice{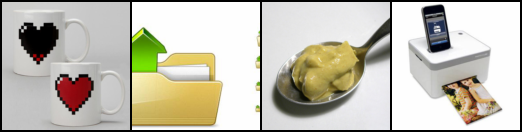}{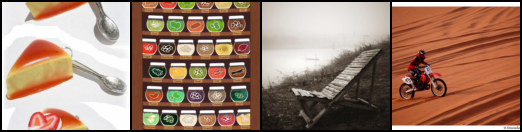}{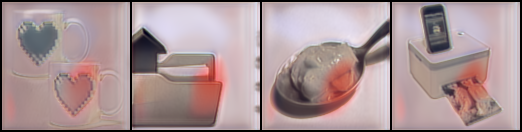}{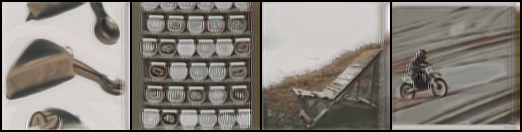}{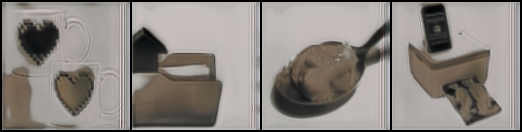}{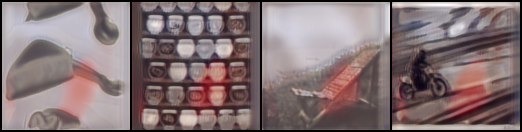} & \genRowOffice{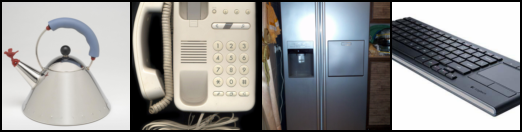}{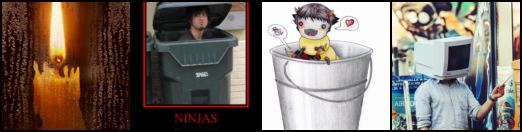}{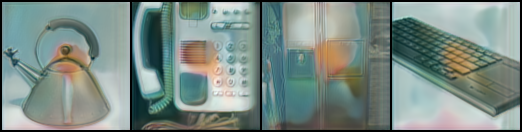}{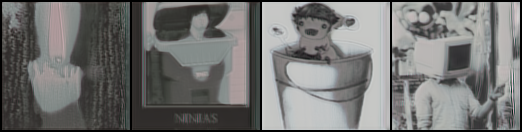}{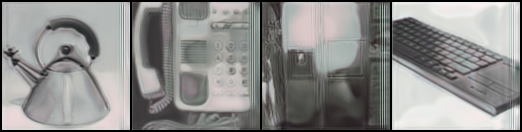}{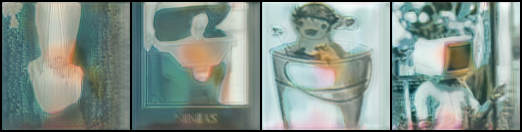} \\ \hline
\textbf{Epoch 20} & \textbf{Epoch 25} \\
\genRowOffice{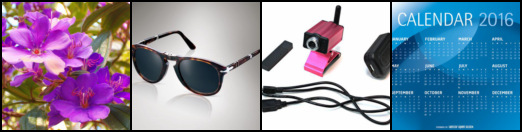}{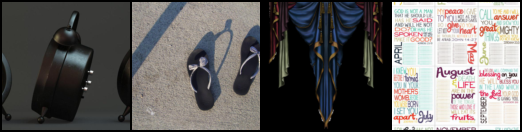}{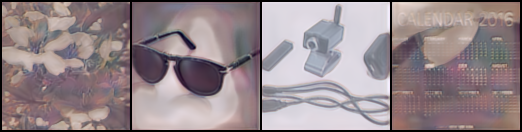}{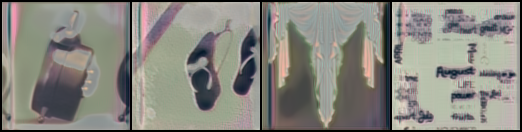}{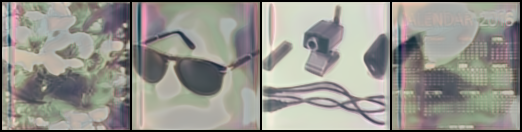}{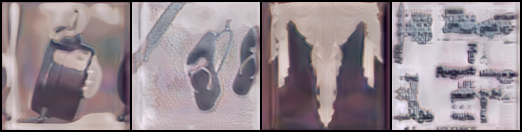} & \genRowOffice{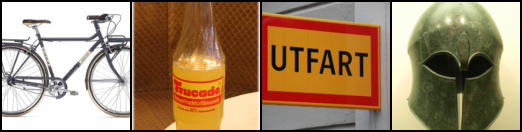}{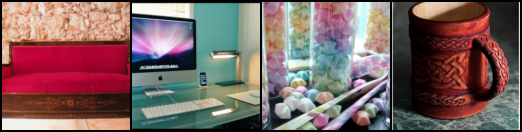}{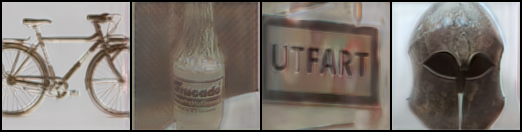}{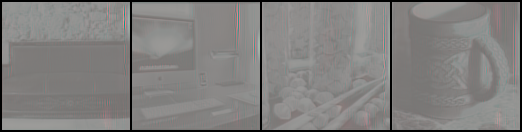}{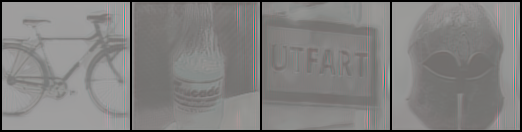}{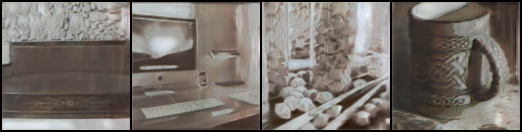} \\ \hline
\end{tabular}
  \caption{The generation results for the
  $\textbf{R} \longleftrightarrow \textbf{Ar}$ transfer task in the \textbf{Office-Home} dataset across epochs $10,15,20,25$. Each quadrant represents results from one epoch, with the one on the left from \textbf{R} domain while on the right is from the \textbf{Ar} domain.}
  \label{table:imge_gen_OneDS}
\end{table*}

\begin{table*}[t]\centering
\begin{tabular}{c|c}
\textbf{Epoch 10} & \textbf{Epoch 15}  \\
\genRowVis{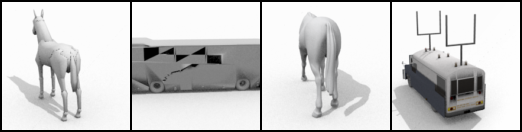}{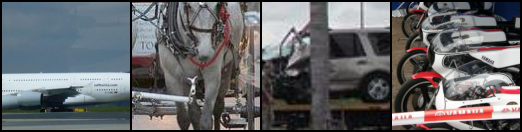}{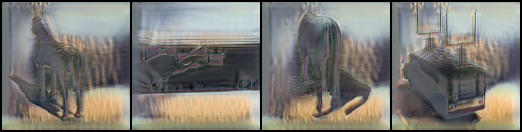}{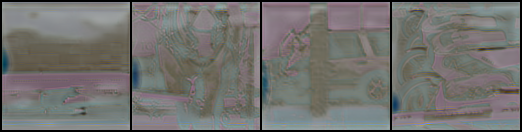}{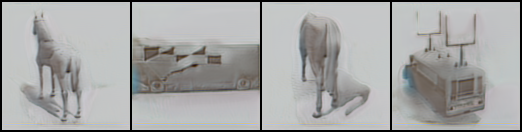}{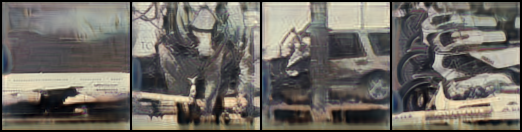} & \genRowVis{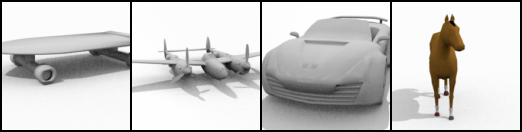}{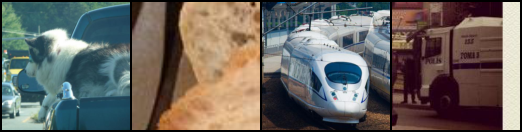}{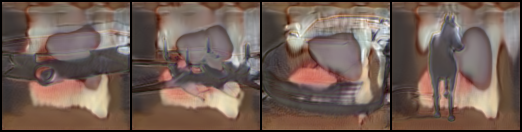}{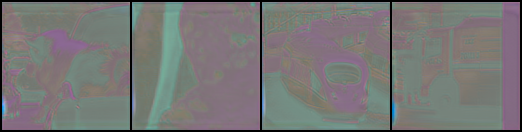}{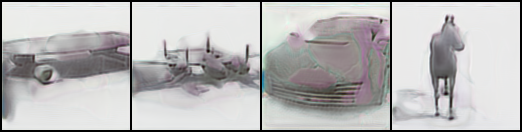}{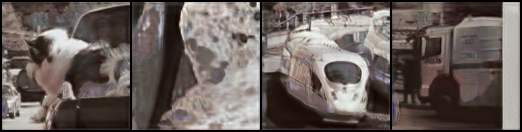} \\ \hline
\textbf{Epoch 20} & \textbf{Epoch 25} \\
\genRowVis{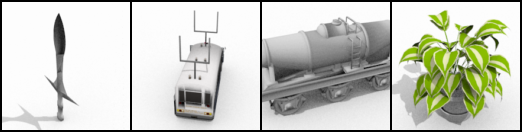}{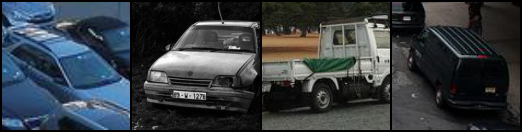}{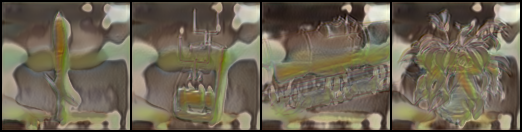}{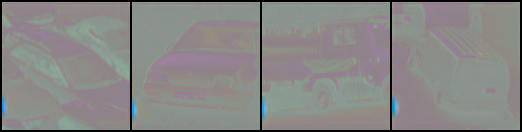}{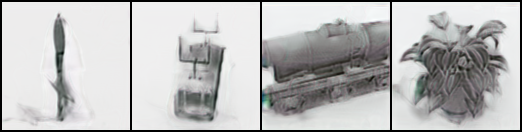}{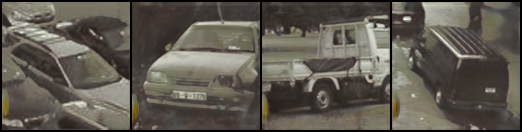} & \genRowVis{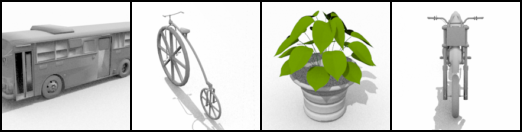}{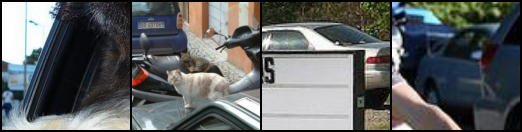}{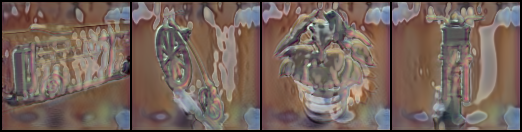}{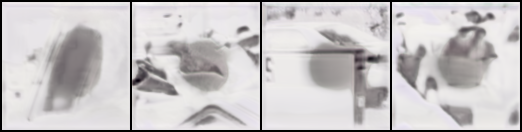}{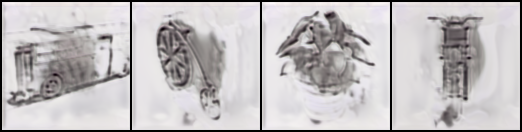}{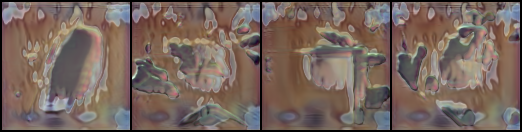} \\ \hline
\end{tabular}
  \vspace{-10.0pt}
  \caption{The generation results for the $\textbf{Sy} \longleftrightarrow \textbf{Re}$  transfer task in the
  \textbf{VisDa} dataset 
    across epochs $10,15,20,25$. In each quadrant, the left block of images represents the synthetic images while the right block represents the real images.}
  \label{table:imge_OneDS_visDA}
\end{table*}

\begin{table*}[t]\centering
\begin{tabular}{c|c}
\textbf{Office-31} $(\textbf{A}mazon \longleftrightarrow \textbf{D}slr)$ & \textbf{Office-31} $(\textbf{A}mazon \longleftrightarrow \textbf{W}eb)$ \\
\genRowAmazon{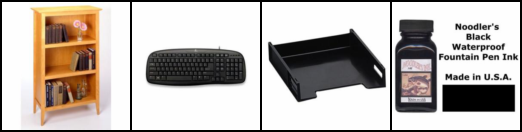}{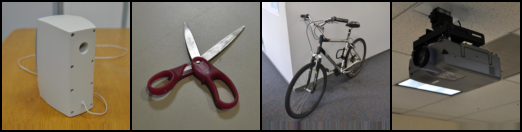}{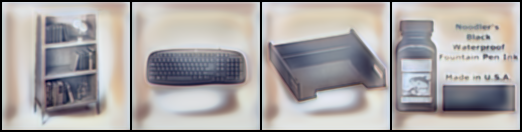}{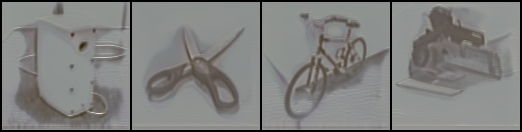}{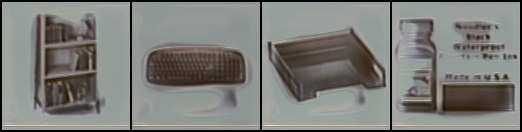}{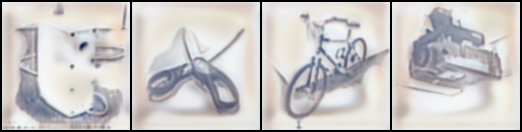}  & \genRowAmazon{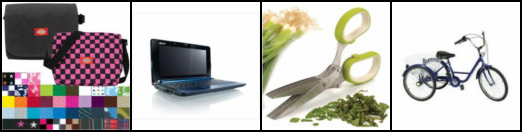}{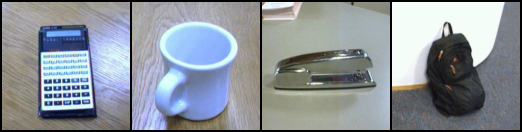}{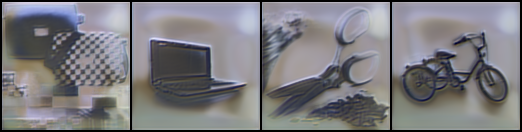}{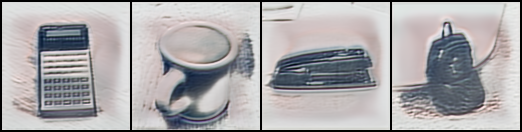}{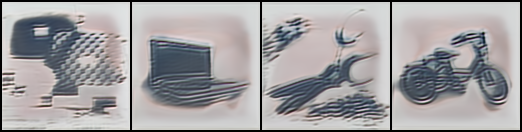}{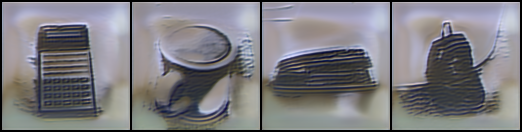}\\ \hline
\textbf{Office-31} $(\textbf{D}slr \longleftrightarrow \textbf{W}eb)$  & \textbf{Office-Home} $(\textbf{Ar}t \longleftrightarrow \textbf{Cl}ip)$  \\
\genRowAmazon{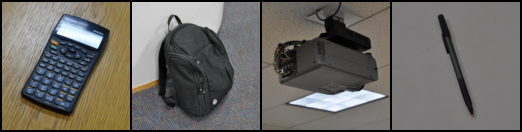}{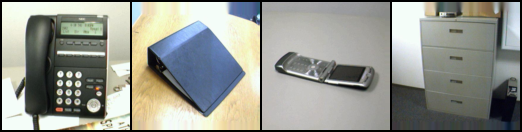}{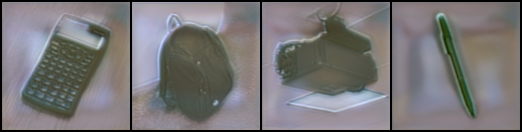}{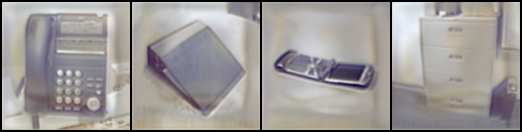}{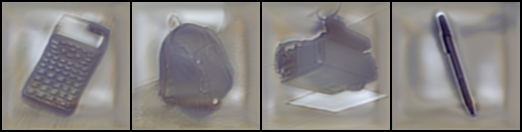}{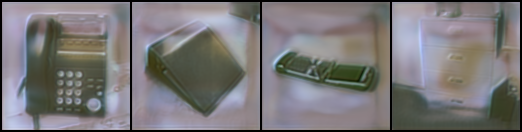} & \genRowOffice{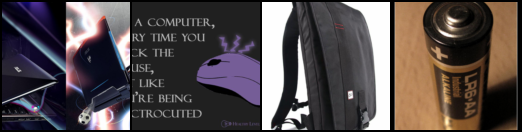}{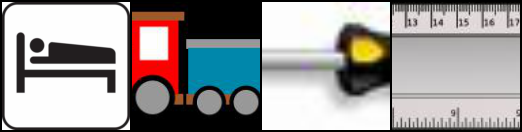}{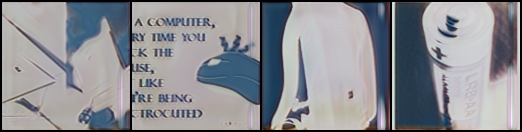}{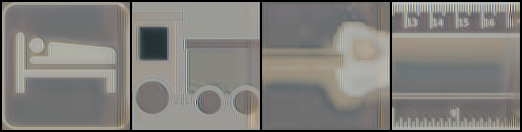}{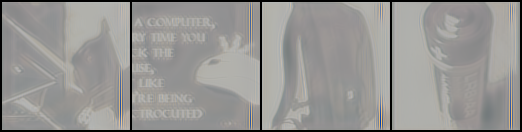}{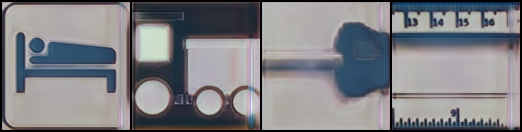} \\ \hline
\textbf{Office-Home} $(\textbf{R}eal \longleftrightarrow \textbf{Ar}t)$ & \textbf{Office-Home} $(\textbf{Ar}t \longleftrightarrow \textbf{P}roduct)$  \\
\genRowOffice{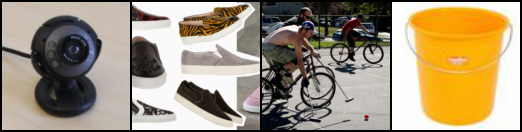}{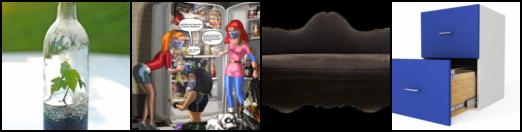}{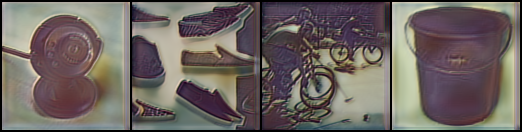}{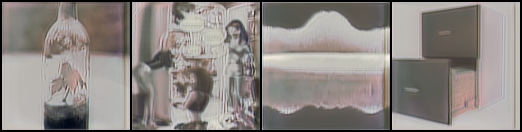}{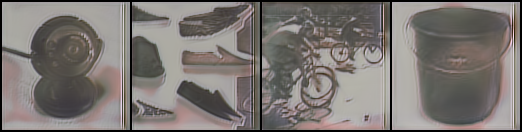}{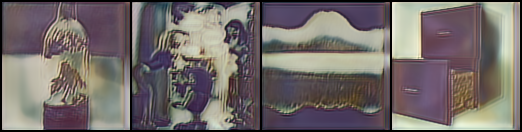} & \genRowOffice{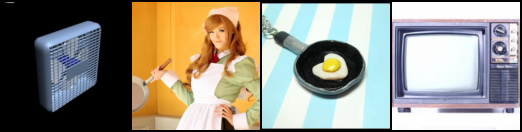}{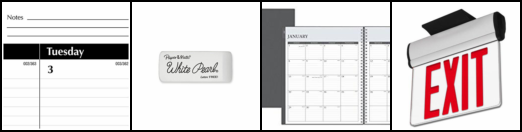}{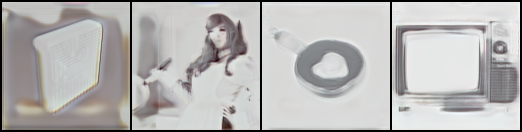}{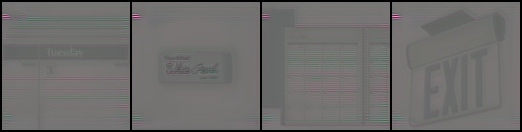}{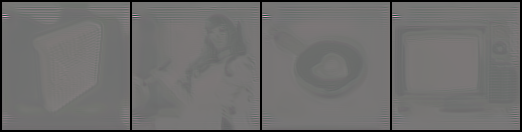}{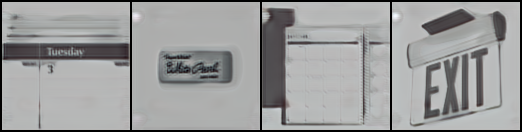} \\ \hline
\textbf{Office-Home} $(\textbf{Cl}ip \longleftrightarrow \textbf{P}roduct)$& \textbf{Office-Home} $(\textbf{Cl}ip \longleftrightarrow \textbf{R}eal)$ \\
\genRowOffice{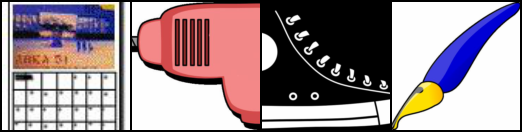}{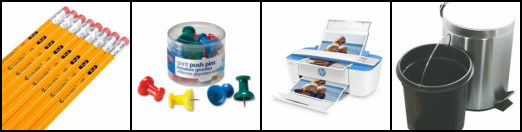}{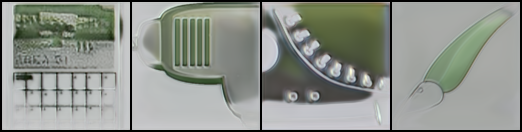}{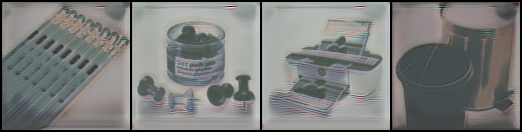}{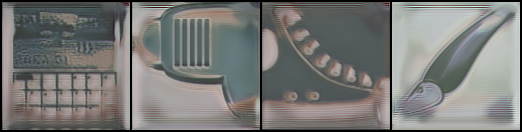}{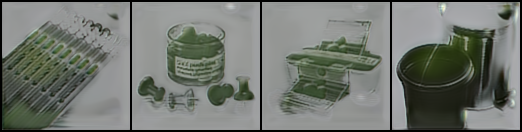} & \genRowOffice{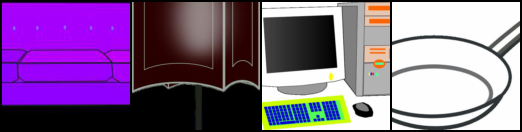}{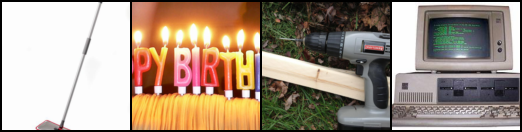}{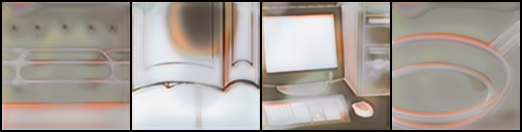}{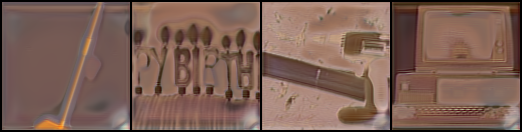}{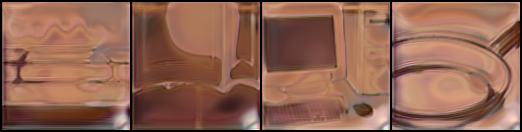}{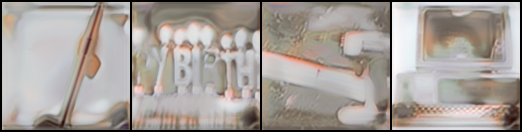} \\ \hline
\textbf{Office-Home} $(\textbf{P}roduct \longleftrightarrow \textbf{R}eal)$ & \textbf{VisDa}  $(\textbf{Sy}nthetic \longleftrightarrow \textbf{R}eal)$  \\
\genRowOffice{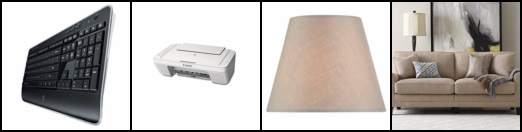}{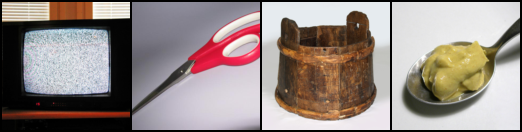}{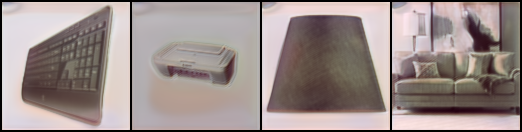}{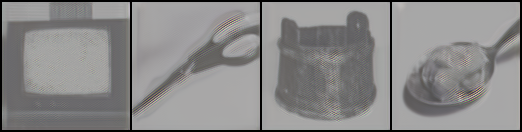}{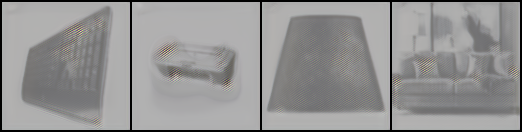}{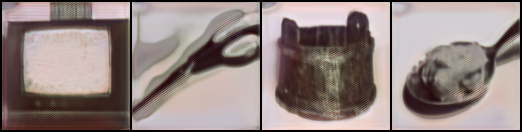} & \genRowVis{name_Unet_basic_visDA__e2e__srcReal_epoch_15.png}{name_Unet_basic_visDA__e2e__srcTarget_epoch_15.png}{name_Unet_basic_visDA__e2e__srcUNET_epoch_15.png}{name_Unet_basic_visDA__e2e__targetUNET_epoch_15.png}{name_Unet_basic_visDA__e2e__src_target_epoch_15.png}{name_Unet_basic_visDA__e2e__target_src_epoch_15.png} \\
\end{tabular}
  \caption{The generation results of epoch 5 across the three datasets and the different transfer tasks. Each cell represents the result of one transfer task.}
  \label{table:imge_gen_All}
\end{table*}

\begin{table*}
\vspace{-20.0pt}
\begin{subtable}[t]{1\textwidth}
\small\addtolength{\tabcolsep}{-5.0pt}
\begin{tabular}{|l|c|c|c|c|c|c|c|c|c|c|c|c|c|c|c|c|c|c|c|c|c|c|c|c|c|c|c|c|c|c|c|c|c|c|c|c|c|c|c|c|c|c|c|c|c|c|c|c|c|c|c|c|c|c|c|c|c|c|c|c|c|c|c|c|c|}\hline 

&\cellcolor{blue!25} M  0& \cellcolor{blue!25} M  1& \cellcolor{blue!25} M  2& \cellcolor{blue!25} M  3& \cellcolor{green!25} M  4& \cellcolor{green!25} M  5& \cellcolor{green!25} M  6& \cellcolor{green!25} M  7& \cellcolor{green!25} M  8& \cellcolor{green!25} M  9& \cellcolor{red!25} M  10& \cellcolor{red!25} M  11& \cellcolor{red!25} M  12& \cellcolor{red!25} M  13& \cellcolor{red!25} M  14& \cellcolor{red!25} M  15\\ \hline 
$D_{Ar}$ & \cellcolor{green!100} 0.8507 &\cellcolor{yellow!100!red!10!blue!10} 0.4898 &\cellcolor{yellow!100!red!10!blue!10} 0.4972 &\cellcolor{yellow!100!red!10!blue!10} 0.5985 &\cellcolor{green!100} 0.7908 &\cellcolor{red!100!green!60} 0.6044 &\cellcolor{green!100} 0.8260 &\cellcolor{yellow!100!red!10!blue!10} 0.5127 &\cellcolor{yellow!100!red!10!blue!10} 0.5076 &\cellcolor{yellow!100!red!10!blue!10} 0.5193 &\cellcolor{green!100} 0.8226 &\cellcolor{red!100!green!60} \textbf{0.7608} &\cellcolor{green!100} 0.8290 &\cellcolor{yellow!100!red!10!blue!10} 0.6745 &\cellcolor{yellow!100!red!10!blue!10} 0.6986 &\cellcolor{yellow!100!red!10!blue!10} 0.7200\\ \hline 
$D_{Ar_{Ar,Cl}}$ &\cellcolor{green!5!yellow} 0.3027 &\cellcolor{yellow!100!red!10!blue!10} 0.2065 &\cellcolor{yellow!100!red!10!blue!10} 0.1979 &\cellcolor{yellow!100!red!10!blue!10} 0.2312 &\cellcolor{green!100} 0.5355 &\cellcolor{yellow!100!red!10!blue!10} 0.2550 &\cellcolor{green!5!yellow} 0.3701 &\cellcolor{yellow!100!red!10!blue!10} 0.2527 &\cellcolor{yellow!100!red!10!blue!10} 0.2496 &\cellcolor{yellow!100!red!10!blue!10} 0.2204 &\cellcolor{green!100} 0.5404 &\cellcolor{yellow!100!red!10!blue!10} \textbf{0.3530} &\cellcolor{green!5!yellow} 0.4220 &\cellcolor{yellow!100!red!10!blue!10} 0.3211 &\cellcolor{yellow!100!red!10!blue!10} 0.3385 &\cellcolor{yellow!100!red!10!blue!10} 0.2954\\ \hline 
$D_{Ar_{Ar,R}}$ &\cellcolor{green!5!yellow} 0.5015 &\cellcolor{yellow!100!red!10!blue!10} 0.2784 &\cellcolor{yellow!100!red!10!blue!10} 0.2879 &\cellcolor{yellow!100!red!10!blue!10} 0.3288 &\cellcolor{green!5!yellow} 0.5424 &\cellcolor{red!100!green!60} 0.3920 &\cellcolor{green!5!yellow} 0.6045 &\cellcolor{yellow!100!red!10!blue!10} 0.3381 &\cellcolor{yellow!100!red!10!blue!10} 0.3299 &\cellcolor{yellow!100!red!10!blue!10} 0.3284 &\cellcolor{green!5!yellow} 0.5851 &\cellcolor{red!100!green!60} \textbf{0.5390} &\cellcolor{green!5!yellow} 0.6336 &\cellcolor{yellow!100!red!10!blue!10} 0.4667 &\cellcolor{yellow!100!red!10!blue!10} 0.4715 &\cellcolor{yellow!100!red!10!blue!10} 0.4876\\ \hline 
$D_{Ar_{Ar,P}}$ &\cellcolor{green!5!yellow} 0.4693 &\cellcolor{yellow!100!red!10!blue!10} 0.2765 &\cellcolor{yellow!100!red!10!blue!10} 0.2907 &\cellcolor{yellow!100!red!10!blue!10} 0.3374 &\cellcolor{green!5!yellow} 0.5186 &\cellcolor{yellow!100!red!10!blue!10} 0.4093 &\cellcolor{green!100} 0.6589 &\cellcolor{yellow!100!red!10!blue!10} 0.3434 &\cellcolor{yellow!100!red!10!blue!10} 0.3380 &\cellcolor{yellow!100!red!10!blue!10} 0.3228 &\cellcolor{green!5!yellow} 0.5819 &\cellcolor{yellow!100!red!10!blue!10} \textbf{0.5186} &\cellcolor{green!100} 0.6496 &\cellcolor{yellow!100!red!10!blue!10} 0.4445 &\cellcolor{yellow!100!red!10!blue!10} 0.4575 &\cellcolor{yellow!100!red!10!blue!10} 0.4477\\ \hline 
$D_{Cl}$  &\cellcolor{yellow!100!red!10!blue!10} 0.5092 &\cellcolor{green!100} 0.8009 &\cellcolor{yellow!100!red!10!blue!10} 0.4856 &\cellcolor{yellow!100!red!10!blue!10} 0.5320 &\cellcolor{red!100!green!60} 0.4706 &\cellcolor{yellow!100!red!10!blue!10} 0.5023 &\cellcolor{yellow!100!red!10!blue!10} 0.4919 &\cellcolor{green!100} 0.7924 &\cellcolor{green!100} 0.7905 &\cellcolor{yellow!100!red!10!blue!10} 0.4790 &\cellcolor{red!100!green!60} 0.5490 &\cellcolor{yellow!100!red!10!blue!10} \textbf{0.5511} &\cellcolor{yellow!100!red!10!blue!10} 0.5376 &\cellcolor{green!100} 0.7410 &\cellcolor{green!100} 0.7036 &\cellcolor{yellow!100!red!10!blue!10} 0.5337\\ \hline 
$D_{Cl_{Cl,Ar}}$ & \cellcolor{yellow!100!red!10!blue!10} 0.3306 &\cellcolor{green!5!yellow} 0.4695 &\cellcolor{yellow!100!red!10!blue!10} 0.2895 &\cellcolor{yellow!100!red!10!blue!10} 0.3262 &\cellcolor{red!100!green!60} 0.3770 &\cellcolor{yellow!100!red!10!blue!10} 0.3542 &\cellcolor{yellow!100!red!10!blue!10} 0.3698 &\cellcolor{green!5!yellow} 0.5546 &\cellcolor{green!5!yellow} 0.5896 &\cellcolor{yellow!100!red!10!blue!10} 0.3308 &\cellcolor{red!100!green!60} 0.4143 &\cellcolor{yellow!100!red!10!blue!10} \textbf{0.4224} &\cellcolor{yellow!100!red!10!blue!10} 0.4029 &\cellcolor{green!5!yellow} 0.6010 &\cellcolor{green!5!yellow} 0.5933 &\cellcolor{yellow!100!red!10!blue!10} 0.3798\\ \hline 
$D_{Cl_{Cl,P}}$ &\cellcolor{yellow!100!red!10!blue!10} 0.3259 &\cellcolor{green!5!yellow} 0.4427 &\cellcolor{yellow!100!red!10!blue!10} 0.2967 &\cellcolor{yellow!100!red!10!blue!10} 0.3264 &\cellcolor{yellow!100!red!10!blue!10} 0.3816 &\cellcolor{yellow!100!red!10!blue!10} 0.3718 &\cellcolor{yellow!100!red!10!blue!10} 0.3907 &\cellcolor{green!100} 0.6592 &\cellcolor{green!5!yellow} 0.5674 &\cellcolor{yellow!100!red!10!blue!10} 0.3349 &\cellcolor{yellow!100!red!10!blue!10} 0.4195 &\cellcolor{yellow!100!red!10!blue!10} \textbf{0.4394} &\cellcolor{yellow!100!red!10!blue!10} 0.4206 &\cellcolor{green!100} 0.6740 &\cellcolor{green!5!yellow} 0.5810 &\cellcolor{yellow!100!red!10!blue!10} 0.3929\\ \hline 
$D_{Cl_{Cl,R}}$  & \cellcolor{yellow!100!red!10!blue!10} 0.3469 &\cellcolor{green!5!yellow} 0.4836 &\cellcolor{yellow!100!red!10!blue!10} 0.3139 &\cellcolor{yellow!100!red!10!blue!10} 0.3328 &\cellcolor{yellow!100!red!10!blue!10} 0.3967 &\cellcolor{yellow!100!red!10!blue!10} 0.3771 &\cellcolor{yellow!100!red!10!blue!10} 0.3914 &\cellcolor{green!5!yellow} 0.5685 &\cellcolor{green!100} 0.6659 &\cellcolor{yellow!100!red!10!blue!10} 0.3382 &\cellcolor{yellow!100!red!10!blue!10} 0.4175 &\cellcolor{yellow!100!red!10!blue!10} \textbf{0.4405} &\cellcolor{yellow!100!red!10!blue!10} 0.3953 &\cellcolor{green!5!yellow} 0.5988 &\cellcolor{green!100} 0.6269 &\cellcolor{yellow!100!red!10!blue!10} 0.3812\\ \hline 
$D_{P}$  &\cellcolor{yellow!100!red!10!blue!10} 0.6261 &\cellcolor{yellow!100!red!10!blue!10} 0.5785 &\cellcolor{green!100} 0.9074 &\cellcolor{yellow!100!red!10!blue!10} \textbf{0.7276} &\cellcolor{yellow!100!red!10!blue!10} 0.5815 &\cellcolor{yellow!100!red!10!blue!10} 0.7103 &\cellcolor{red!100!green!60} 0.6227 &\cellcolor{red!100!green!60} 0.6012 &\cellcolor{yellow!100!red!10!blue!10} 0.6128 &\cellcolor{green!100} 0.9090 &\cellcolor{yellow!100!red!10!blue!10} 0.6493 &\cellcolor{yellow!100!red!10!blue!10} 0.7017 &\cellcolor{red!100!green!60} 0.6589 &\cellcolor{red!100!green!60} 0.6383 &\cellcolor{yellow!100!red!10!blue!10} 0.6335 &\cellcolor{green!100} 0.8370\\ \hline 
$D_{P_{Ar,P}}$ &\cellcolor{yellow!100!red!10!blue!10} 0.3173 &\cellcolor{yellow!100!red!10!blue!10} 0.2529 &\cellcolor{green!5!yellow} 0.4147 &\cellcolor{yellow!100!red!10!blue!10} 0.3112 &\cellcolor{yellow!100!red!10!blue!10} 0.3110 &\cellcolor{yellow!100!red!10!blue!10} 0.3587 &\cellcolor{red!100!green!60} 0.3501 &\cellcolor{yellow!100!red!10!blue!10} 0.3522 &\cellcolor{yellow!100!red!10!blue!10} 0.3106 &\cellcolor{green!5!yellow} 0.4667 &\cellcolor{yellow!100!red!10!blue!10} 0.3450 &\cellcolor{yellow!100!red!10!blue!10} \textbf{0.4322} &\cellcolor{red!100!green!60} 0.3768 &\cellcolor{yellow!100!red!10!blue!10} 0.4043 &\cellcolor{yellow!100!red!10!blue!10} 0.3964 &\cellcolor{green!5!yellow} 0.5538\\ \hline 
$D_{P_{Cl,P}}$ &\cellcolor{yellow!100!red!10!blue!10} 0.4256 &\cellcolor{yellow!100!red!10!blue!10} 0.3695 &\cellcolor{green!5!yellow} 0.5605 &\cellcolor{yellow!100!red!10!blue!10} 0.4486 &\cellcolor{yellow!100!red!10!blue!10} 0.4218 &\cellcolor{yellow!100!red!10!blue!10} 0.5035 &\cellcolor{yellow!100!red!10!blue!10} 0.4424 &\cellcolor{red!100!green!60} 0.4291 &\cellcolor{yellow!100!red!10!blue!10} 0.4238 &\cellcolor{green!5!yellow} 0.6832 &\cellcolor{yellow!100!red!10!blue!10} 0.4633 &\cellcolor{yellow!100!red!10!blue!10} \textbf{0.5534} &\cellcolor{yellow!100!red!10!blue!10} 0.4965 &\cellcolor{red!100!green!60} 0.4980 &\cellcolor{yellow!100!red!10!blue!10} 0.4945 &\cellcolor{green!5!yellow} 0.7168\\ \hline 
$D_{P_{P,R}}$ &\cellcolor{yellow!100!red!10!blue!10} 0.4885 &\cellcolor{yellow!100!red!10!blue!10} 0.4229 &\cellcolor{green!5!yellow} 0.6851 &\cellcolor{yellow!100!red!10!blue!10} 0.5363 &\cellcolor{yellow!100!red!10!blue!10} 0.4638 &\cellcolor{yellow!100!red!10!blue!10} 0.5785 &\cellcolor{yellow!100!red!10!blue!10} 0.5029 &\cellcolor{yellow!100!red!10!blue!10} 0.4745 &\cellcolor{yellow!100!red!10!blue!10} 0.4885 &\cellcolor{green!100} 0.8371 &\cellcolor{yellow!100!red!10!blue!10} 0.5223 &\cellcolor{yellow!100!red!10!blue!10} \textbf{0.6164} &\cellcolor{yellow!100!red!10!blue!10} 0.5414 &\cellcolor{yellow!100!red!10!blue!10} 0.5413 &\cellcolor{yellow!100!red!10!blue!10} 0.5403 &\cellcolor{green!100} 0.7950\\ \hline 
$D_{R}$ &\cellcolor{yellow!100!red!10!blue!10} 0.7113 &\cellcolor{yellow!100!red!10!blue!10} 0.6296 &\cellcolor{yellow!100!red!10!blue!10} 0.6998 &\cellcolor{green!100} 0.8964 &\cellcolor{yellow!100!red!10!blue!10} 0.6688 &\cellcolor{green!100} 0.8737 &\cellcolor{yellow!100!red!10!blue!10} 0.7020 &\cellcolor{yellow!100!red!10!blue!10} 0.6472 &\cellcolor{red!100!green!60} 0.6403 &\cellcolor{red!100!green!60} 0.7188 &\cellcolor{yellow!100!red!10!blue!10} 0.6975 &\cellcolor{green!100} 0.7786 &\cellcolor{yellow!100!red!10!blue!10} 0.7205 &\cellcolor{yellow!100!red!10!blue!10} 0.6830 &\cellcolor{red!100!green!60} 0.6806 &\cellcolor{red!100!green!60} \textbf{0.7537}\\ \hline 
$D_{R_{Ar,R}}$ &\cellcolor{yellow!100!red!10!blue!10} 0.4773 &\cellcolor{yellow!100!red!10!blue!10} 0.3766 &\cellcolor{yellow!100!red!10!blue!10} 0.4036 &\cellcolor{green!5!yellow} 0.5354 &\cellcolor{yellow!100!red!10!blue!10} 0.4847 &\cellcolor{green!100} 0.7461 &\cellcolor{yellow!100!red!10!blue!10} 0.5348 &\cellcolor{yellow!100!red!10!blue!10} 0.4476 &\cellcolor{yellow!100!red!10!blue!10} 0.4399 &\cellcolor{yellow!100!red!10!blue!10} 0.4606 &\cellcolor{yellow!100!red!10!blue!10} 0.5084 &\cellcolor{green!100} 0.6700 &\cellcolor{yellow!100!red!10!blue!10} \textbf{0.5436} &\cellcolor{yellow!100!red!10!blue!10} 0.5221 &\cellcolor{yellow!100!red!10!blue!10} 0.5310 &\cellcolor{yellow!100!red!10!blue!10} 0.5326\\ \hline 
$D_{R_{Cl,R}}$&\cellcolor{yellow!100!red!10!blue!10} 0.2900 &\cellcolor{yellow!100!red!10!blue!10} 0.2769 &\cellcolor{yellow!100!red!10!blue!10} 0.2821 &\cellcolor{green!5!yellow} 0.3278 &\cellcolor{yellow!100!red!10!blue!10} 0.3584 &\cellcolor{green!5!yellow} 0.3876 &\cellcolor{yellow!100!red!10!blue!10} 0.3398 &\cellcolor{yellow!100!red!10!blue!10} 0.3168 &\cellcolor{red!100!green!60} 0.3197 &\cellcolor{yellow!100!red!10!blue!10} 0.3013 &\cellcolor{yellow!100!red!10!blue!10} 0.3861 &\cellcolor{green!5!yellow} 0.4339 &\cellcolor{yellow!100!red!10!blue!10} 0.3550 &\cellcolor{yellow!100!red!10!blue!10} 0.3853 &\cellcolor{red!100!green!60} \textbf{0.3987} &\cellcolor{yellow!100!red!10!blue!10} 0.3473\\ \hline 
$D_{R_{P,R}}$ &\cellcolor{yellow!100!red!10!blue!10} 0.5084 &\cellcolor{yellow!100!red!10!blue!10} 0.4024 &\cellcolor{yellow!100!red!10!blue!10} 0.4574 &\cellcolor{green!5!yellow} 0.6071 &\cellcolor{yellow!100!red!10!blue!10} 0.4970 &\cellcolor{green!5!yellow} 0.6863 &\cellcolor{yellow!100!red!10!blue!10} 0.5355 &\cellcolor{yellow!100!red!10!blue!10} 0.4745 &\cellcolor{yellow!100!red!10!blue!10} 0.4638 &\cellcolor{red!100!green!60} 0.5079 &\cellcolor{yellow!100!red!10!blue!10} 0.5363 &\cellcolor{green!5!yellow} 0.6458 &\cellcolor{yellow!100!red!10!blue!10} 0.5647 &\cellcolor{yellow!100!red!10!blue!10} 0.5442 &\cellcolor{yellow!100!red!10!blue!10} 0.5480 &\cellcolor{red!100!green!60} \textbf{0.5885}\\ \hline 
\end{tabular}
\caption{Top-5 classification results for \textbf{Office-Home} dataset,  showing that model \textbf{M 11} achieves the best performance across all other transfer tasks.}
\label{table:res_office}
\end{subtable}
\vspace{-80.0pt}
\begin{subtable}[t]{0.535\textwidth}
\small\addtolength{\tabcolsep}{-5.0pt}
\begin{tabular}{|l|c|c|c|c|c|c|c|c|c|c|c|c|c|c|c|c|c|c|c|c|c|c|c|c|c|c|c|c|c|c|c|c|c|}\hline 
 & \cellcolor{blue!25} M  0& \cellcolor{blue!25} M  1 & \cellcolor{blue!25} M  2& \cellcolor{green!25} M  3& \cellcolor{green!25} M  4& \cellcolor{green!25} M  5& \cellcolor{red!25} M  6& \cellcolor{red!25} M  7& \cellcolor{red!25} M  8\\ \hline 
$D_{A}$ &\cellcolor{green!100} 0.904 & \cellcolor{yellow!100!red!10!blue!10} 0.006 & \cellcolor{yellow!100!red!10!blue!10} 0.0038 &\cellcolor{green!100} 0.901 &\cellcolor{green!100} 0.885 &\cellcolor{yellow!100!red!10!blue!10} 0.622 &\cellcolor{green!100} 0.959 &\cellcolor{green!100} 0.955 &\cellcolor{yellow!100!red!10!blue!10} \textbf{0.868}\\ \hline 
$D_{A_{A,D}}$ &\cellcolor{green!5!yellow} 0.551 & \cellcolor{yellow!100!red!10!blue!10} 0.012 & \cellcolor{yellow!100!red!10!blue!10} 0.0012 &\cellcolor{green!100} 0.806 &\cellcolor{green!5!yellow} 0.742 &\cellcolor{yellow!100!red!10!blue!10} 0.548 &\cellcolor{green!100} 0.866 &\cellcolor{green!5!yellow} 0.823 &\cellcolor{yellow!100!red!10!blue!10} \textbf{0.700}\\ \hline 
$D_{A_{A,W}}$ &\cellcolor{green!5!yellow} 0.578 & \cellcolor{yellow!100!red!10!blue!10} 0.029 & \cellcolor{yellow!100!red!10!blue!10} 0.00 & \cellcolor{green!5!yellow} 0.721 &\cellcolor{green!100} 0.790 &\cellcolor{yellow!100!red!10!blue!10} 0.545 &\cellcolor{green!5!yellow} 0.834 &\cellcolor{green!100} 0.857 &\cellcolor{yellow!100!red!10!blue!10} \textbf{0.690}\\ \hline 
$D_{D}$ &\cellcolor{yellow!100!red!10!blue!10} 0.728 & \cellcolor{green!100} 0.650 & \cellcolor{yellow!100!red!10!blue!10} 0.549 & \cellcolor{red!100!green!60} 0.775 &\cellcolor{yellow!100!red!10!blue!10} 0.795 &\cellcolor{green!100} 0.973 &\cellcolor{red!100!green!60} 0.855 &\cellcolor{yellow!100!red!10!blue!10} \textbf{0.865} &\cellcolor{green!100} 0.994\\ \hline 
$D_{D_{A,D}}$ &\cellcolor{yellow!100!red!10!blue!10} 0.482 & \cellcolor{green!5!yellow} 0.136 & \cellcolor{yellow!100!red!10!blue!10} 0.156 & \cellcolor{red!100!green!60} 0.587 &\cellcolor{yellow!100!red!10!blue!10} 0.631 &\cellcolor{green!5!yellow} 0.853 &\cellcolor{red!100!green!60} 0.690 &\cellcolor{yellow!100!red!10!blue!10} \textbf{0.696} &\cellcolor{green!5!yellow} 0.901\\ \hline 
$D_{D_{W,D}}$ &\cellcolor{yellow!100!red!10!blue!10} 0.418 & \cellcolor{green!5!yellow} 0.091 & \cellcolor{yellow!100!red!10!blue!10} 0.169 & \cellcolor{yellow!100!red!10!blue!10} 0.524 &\cellcolor{yellow!100!red!10!blue!10} 0.592 &\cellcolor{green!100} 0.899 &\cellcolor{yellow!100!red!10!blue!10} 0.634 &\cellcolor{yellow!100!red!10!blue!10} \textbf{0.673} &\cellcolor{green!100} 0.968\\ \hline 
$D_{W}$  &\cellcolor{yellow!100!red!10!blue!10} 0.692 & \cellcolor{yellow!100!red!10!blue!10} 0.34 & \cellcolor{green!100} 0.553 &\cellcolor{yellow!100!red!10!blue!10} 0.750 &\cellcolor{red!100!green!60} 0.754 &\cellcolor{red!100!green!60} 0.934 &\cellcolor{yellow!100!red!10!blue!10} 0.832 &\cellcolor{red!100!green!60} 0.838 &\cellcolor{red!100!green!60} \textbf{0.979}\\ \hline 
$D_{W_{W,A}}$  &\cellcolor{yellow!100!red!10!blue!10} 0.429 & \cellcolor{yellow!100!red!10!blue!10} 0.040 & \cellcolor{green!5!yellow} 0.056 &\cellcolor{yellow!100!red!10!blue!10} 0.517 &\cellcolor{red!100!green!60} 0.560 &\cellcolor{yellow!100!red!10!blue!10} 0.801 &\cellcolor{yellow!100!red!10!blue!10} 0.600 &\cellcolor{red!100!green!60} 0.624 &\cellcolor{yellow!100!red!10!blue!10} \textbf{0.823}\\ \hline 
$D_{W_{W,D}}$  &\cellcolor{yellow!100!red!10!blue!10} 0.472 & \cellcolor{yellow!100!red!10!blue!10} 0.068 & \cellcolor{green!5!yellow} 0.139 &\cellcolor{yellow!100!red!10!blue!10} 0.591 &\cellcolor{yellow!100!red!10!blue!10} 0.601 &\cellcolor{red!100!green!60} 0.858 &\cellcolor{yellow!100!red!10!blue!10} 0.677 &\cellcolor{yellow!100!red!10!blue!10} 0.691 &\cellcolor{red!100!green!60} \textbf{0.918}\\ \hline 
\end{tabular}
\caption{Top-5 classification results for \textbf{Office-31} dataset, showing that    
model \textbf{M 8} achieves the best performance across all other transfer tasks.}
\label{table:res_amazon}
\end{subtable}
\vspace{-60.0pt}
\hspace{0.0pt}
\begin{subtable}[t]{0.45\textwidth}
\small\addtolength{\tabcolsep}{-6.5pt}
\vspace{-10.0pt}
\begin{tabular}{|c|c|c|c|c|}\hline 
 & \cellcolor{blue!25} M  0 & \cellcolor{green!25} M  1 & \cellcolor{red!25} M 2 &\cellcolor{red!25} M 3 \\ \hline
$D_{Sy}$ & \cellcolor{green!100} 0.954 (0.799) & \cellcolor{green!100} 0.934 (0.789) & \cellcolor{green!100} 0.909 (0.777) & \cellcolor{green!100} 0.916 (0.786) \\ \hline
$D_{Sy_{Sy,Re}}$ & \cellcolor{green!5!yellow} 0.753 (0.270) & \cellcolor{green!100} 0.948 (0.747) & \cellcolor{green!100} 0.889 (0.746) & \cellcolor{green!100} 0.892 (0.742) \\ \hline
$D_{Re}$ & \cellcolor{red!100!green!60} 0.605 (0.183) & \cellcolor{red!100!green!60} \textbf{0.731} (0.387) & \cellcolor{red!100!green!60} 0.700 (\textbf{0.395}) & \cellcolor{red!100!green!60} 0.705 (0.394) \\ \hline
$D_{Re_{Re,Sy}}$ & \cellcolor{red!100!green!60} 0.511 (0.142)& \cellcolor{red!100!green!60} \textbf{0.577} (0.192) & \cellcolor{red!100!green!60} 0.553 (0.194) & \cellcolor{red!100!green!60} 0.559 (\textbf{0.195}) \\ \hline
\end{tabular}
\caption{Top-5 (Top-1) Classification performance results for \textbf{VisDa}.}
\label{table:res_visDA}
\end{subtable}
\vspace{-20.0pt}
\flushright
\begin{subtable}[t]{0.45\textwidth}
\small\addtolength{\tabcolsep}{-2.5pt}
\begin{tabular}{|c|c|}\hline 
\cellcolor{blue!25} $M_{base}$ & \cellcolor{green!100} $Data_{T1}$ \\ \hline
\cellcolor{green!25} $M_{tune}$ & \cellcolor{green!5!yellow} $Data_{T2}$\\ \hline
\cellcolor{red!25} $M_{ensemb}$ & \cellcolor{red!100!green!60} $Data_{T3}$  \\ \hline
&  \cellcolor{yellow!100!red!10!blue!10}  $Data_{T4}$  \\ \hline
\end{tabular}
\vspace{-5.0pt}
\caption{Color coding of the results for both models and datasets.}
\label{table:model_codes}
\end{subtable}
\vspace{75.0pt}
\label{tab:res_table_all}
\end{table*}

\textbf{CycleGAN-based Stochastic Style Transfer.} 
The results of the Stochastic Style Transfer generation module are shown in Tables \ref{table:imge_gen_OneDS}, \ref{table:imge_OneDS_visDA}, and \ref{table:imge_gen_All}, where Table \ref{table:imge_gen_All} shows the results across multiple transfer tasks in all three datasets, while Tables \ref{table:imge_OneDS_visDA} and \ref{table:imge_gen_All} show results across different training epochs in the same transfer task across different training epochs. These results are shown in a Table format where each cell in the table contains two blocks of images. Each block consists of three rows, the first row represents the original images $I_s$ or $I_t$ for source and target domains respectively (source on the left, target on the right). The second row represents the output of applying the generator network to the input image denoted as $I_{su}$ or $I_{tu}$ for source and target images respectively. The third row represents the cycle reconstructed images $I_{srec}$ or $I_{trec}$. 

For training the module we used a batch size of $8$, and we used the whole dataset in case of \textbf{Office-31}, but in the case of \textbf{Office-Home} we use $85\%$ and in case of \textbf{VisDa} source we used $0.5\%$ and $5\%$ for the target. We did not sample the overall dataset in the cases of \textbf{Office-Home} and \textbf{VisDa} in order to reduce training time due to very small batch size in training. We choose $\lambda_i = 1$ in Eq \ref{eq:7} for $i = 1 \dots 6$

The results show that the module manages to create different image instances of the original domain that have style similar to  the images of the other domain. Table \ref{table:imge_gen_OneDS} shows the transfer results for the Office-Home dataset between two domains $\textbf{R}\longrightarrow \textbf{Ar}$ across different training epochs. It illustrates that the system managed to produce photo realistic results across many epochs except for epoch 25, where the results are little bit washed out. This demonstrates that with different statuses of the module parameters $U_s$ and $U_t$ we can obtain different style transfer results. The figures show that the module managed to keep the contents of the original domain but transfer the style of the other domain in a batch based mode. The results for the VisDa dataset are shown in Table \ref{table:imge_OneDS_visDA}. The performance is less compelling compared to the other dataset as it is a more challenging task. Still the system managed to produce good transfer results between the two domain as in Epoch 10 and Epoch 25. Finally results across all transfer tasks for all the datasets in Epoch 5 of the training algorithm are shown in Table \ref{table:imge_gen_All}. These results illustrate that the system managed to provide good results except for some cases as in \textbf{Office-Home} $(\textbf{Ar}t \longleftrightarrow \textbf{P}roduct)$ and \textbf{VisDa}  $(\textbf{Sy}nthetic \longleftrightarrow \textbf{R}eal)$.

For each domain's data we can generate many different instances of that data using the generation modules trained for different transfer tasks. For example consider the \textbf{A}mazon domain in the \textbf{Office-31} dataset, which has its original version $D_{A}$ and the version generated due to generation module $(\textbf{A}mazon \longleftrightarrow \textbf{D}slr)$ known as $D_{A_{A,D}}$  or due to the module $(\textbf{A}mazon \longleftrightarrow \textbf{W}eb)$ known as $D_{A_{A,W}}$. In each generation module we can choose training parameters stored from different training epochs (in our case we used epoch 5) and we can choose to get the images generated to be $(I_{su},I_{tu})$ or the reconstructed images $(I_{srec},I_{trec})$ (in our case we used $(I_{su},I_{tu})$). 

\textbf{Self-ensemble zero shot domain adaptation.}
The classification results are shown in Tables \ref{table:res_amazon}, \ref{table:res_office}, and \ref{table:res_visDA} for datasets \textbf{Office-31}, \textbf{Office-Home}, and \textbf{VisDa} datasets, respectively. The color coding for both models and the datasets are shown in Table \ref{table:model_codes}. In our experiments we trained three different types of classification models with color codes in Table \ref{table:model_codes}: 1) Base model $M_{base}$: a model that has been trained on original source images only in a supervised way, the source datasets have color described by $Data_{T1}$ in Table \ref{table:model_codes}. 2) Tuned model $M_{tune}$: for a model that has been trained in a supervised way using the original source images and the source mapped images (simple data augmentation to the source), the source datasets both have color described by $Data_{T1}$ in Table \ref{table:model_codes}. 3) $M_{ensemb}$ model: is a model trained in a supervised way using the data described by the source datasets (with color $Data_{T1}$) while trained in the unsupervised way using the target datasets with color described by $Data_{T3}$. Finally, all these models have been tested against other datasets that have not been involved in training, either supervised or unsupervised, with color described by $Data_{T4}$. 

\textbf{Results Analysis.} The performance results show that there is a model that has the highest performance  on all other target domains. Table \ref{table:res_amazon} presents the results of the \textbf{Office-31} dataset showing that Model M $8$ has the highest performance across all the other transfer tasks. This model has been trained using the proposed method with supervised source data $D_{D}$ and $D_{D_{W,D}}$ while trained on $D_{W}$ and $D_{W_{W,D}}$ in an unsupervised way. Despite that this model has never seen any instance of domain $A$, it has the highest transfer performance across all instances of domain $A$. Similarly, the results of the \textbf{Office-Home} dataset are shown  in Table \ref{table:res_office}. From this table, we can see that model M $11$ has the highest performance over all other target datasets except one case of data $D_{P}$. This model was trained on $D_R$ and $D_{R_{Ar,R}}$ in a supervised way while it was trained on $D_{Ar}$ and $D_{Ar_{Ar,R}}$ in an unsupervised way, and managed to generalize across other domains of $P$,$Cl$ without seeing these data. These results are consistent with Top-1 results. Finally, the results of \textbf{VisDA} are presented in Table \ref{table:res_visDA}, showing that model M 1 achieves the best transfer top-5 results, while on top-1, M 2 and M 3 achieve best results on $D_{Re}$ and $D_{Re_{Re,Sy}}$ respectively. No single model managed to get higher performance, we believe this is due to less quality of Generated random instances shown in Table \ref{table:imge_OneDS_visDA}

\section{Conclusion and Future Work}
In this work, we showed that domain adaptation can benefit from generative models to enhance network generalization performance across multiple other domains. We demonstrated that these networks can produce multiple random instances of the same domain dataset. Each instance represents a random different shift compared to the original dataset. Our results also show that using a self-ensemble method is better than simple data augmentation to enhance knowledge transfer performance even on unseen domains. In the future we plan to train the model end-to-end where each epoch of the generation module produces data with new random domain shifts and train the classifier incrementally, so that the model can generalize across continuous domain shift changes. Moreover, we plan to conduct experiments in which we compare the highest performing model across different methods in different transfer tasks in the same dataset, which we did not do here because our goal was to show that a model can generalize even across unseen domains by using the generative models, which to our knowledge has not been done before.

{\small
\bibliographystyle{ieee}
\bibliography{5c0355fe9db46e4e3c1d1a1a_166/egpaper_final}
}

\end{document}